\documentclass[letterpaper]{article}

\usepackage[nonatbib, preprint]{neurips_2023}

\usepackage[T1]{fontenc}
\usepackage{graphicx}
\usepackage{amsmath} 
\usepackage{amssymb}  
\usepackage[style=lncs, maxbibnames=100, citestyle=numeric-comp, sorting=none]{biblatex}
\usepackage{csquotes}
\usepackage{algorithm}
\usepackage{algorithmic}
\usepackage{wrapfig}
\usepackage{hyperref}
\usepackage{subcaption}
\usepackage{forest}
\usepackage{authblk}

\bibliography{references}

\title{\LARGE \bf
Model-based Policy Optimization using Symbolic World Model
}

\author[1,3]{Andrey Gorodetskiy}
\author[1,2]{Konstantin Mironov}
\author[3,2,1]{Aleksandr Panov}

\affil[1]{Center of Cognitive Modeling, Moscow Institute of Physics and Technology}
\affil[2]{Artificial Intelligence Research Institute}
\affil[3]{Federal Research Center ``Computer Science and Control''}

\begin{document}
\maketitle

\begin{abstract}
The application of learning-based control methods in robotics presents significant challenges. One is that model-free reinforcement learning algorithms use observation data with low sample efficiency. To address this challenge, a prevalent approach is model-based reinforcement learning, which involves employing an environment dynamics model. We suggest approximating transition dynamics with symbolic expressions, which are generated via symbolic regression. Approximation of a mechanical system with a symbolic model has fewer parameters than approximation with neural networks, which can potentially lead to higher accuracy and quality of extrapolation. We use a symbolic dynamics model to generate trajectories in model-based policy optimization to improve the sample efficiency of the learning algorithm. We evaluate our approach across various tasks within simulated environments. Our method demonstrates superior sample efficiency in these tasks compared to model-free and model-based baseline methods.
\end{abstract}

\section{Introduction}
Learning-based methods have wide applications in robotics due to the complexity of problems encountered in real-world scenarios. For instance, manipulators often engage with diverse objects under uncertain conditions in non-deterministic environments with unknown parameters. Reinforcement learning (RL) offers a solution by finding effective control policies from interaction data, enabling robots to navigate and manipulate their environments with greater efficiency \parencite{kalashnikov2018qt, kalashnikov2018scalable}.
Reinforcement learning has been successfully applied to many fields, like games \parencite{silver2016mastering, berner2019dota, hafner2023mastering}, natural language processing \parencite{ziegler2019finetuning}, and robotic manipulation \parencite{kalashnikov2018qt, kalashnikov2018scalable}.

In robotic manipulation, RL algorithms have been employed to enable robots to learn dexterous manipulation skills that generalize across different objects and scenarios, such as grasping, picking, and placing objects in diverse environments \cite{gupta2021reset}.
RL can produce policies for robot navigation \cite{staroverov2020real} and path planning \cite{skrynnik2023switch} as well. Autonomous robots were shown to learn to navigate through cluttered environments, avoid obstacles, and reach desired destinations efficiently.
In this work, we aim to overcome the challenge of learning a model of the environment for a model-based RL algorithm by employing the symbolic regression method.
\begin{figure}
    \centering
    \includegraphics[width=0.8\textwidth]{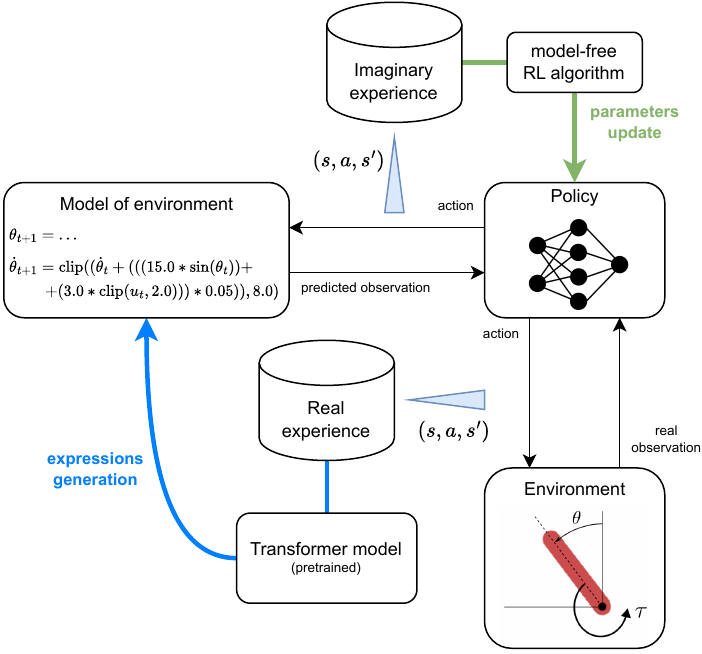}
    \caption{The scheme of model-based policy optimization with symbolic model. We train a policy on samples from an environment model represented by a collection of symbolic expressions. These expressions are generated by a transformer model using observed transitions from the environment. The transformer model is pre-trained on a diverse dataset of randomly generated or environment-specific transition functions.}
    \label{fig:visual-abstract}
\end{figure}
Symbolic regression is a method that discovers a mathematical expression that represents the relationship between input and output variables in a dataset. Unlike traditional regression methods that rely on predefined functional forms, like linear regression or neural networks, symbolic regression aims to find symbolic expressions not limited to specific mathematical structures.
The symbolic regression method involves searching through a space of mathematical expressions to find the one that best fits the given data. This search is typically performed using optimization techniques such as genetic algorithms or evolutionary algorithms. These algorithms iteratively generate and evaluate candidate expressions, adjusting them over successive iterations to improve their fit to the data.
However, symbolic regression can be computationally expensive, especially for large datasets or complex expressions.
Recent advances in deep learning led to a new method of solving symbolic regression problems using transformer architecture to generate symbolic expressions given a dataset \parencite{kamienny2022end}. This method showed performance similar to state-of-the-art genetic programming methods, required less inference time, and expanded the applicability to higher input dimensions, thus motivating our interest in utilizing it to solve model-based reinforcement learning (MBRL) problems.

There are several potential advantages of using the symbolic regression method and symbolic expressions for modeling the environment dynamics in model-based policy optimization \parencite{ha2018world, hafner2019learning, yildiz2021continuous}.
First, it can provide a much simpler model of dynamics in the desired region of state space, with a higher quality of extrapolation for neighboring regions of state space, since the model can capture the exact relationship rather than just numerically approximating it.
Second, the simplicity of symbolic expression leads to lower computational complexity, which gives the model-based algorithm more computational budget.
This suggests it may be a promising approach for modeling environment dynamics in MBRL algorithms.

We propose a new method for policy learning in which a model-free RL algorithm is trained on samples from a world model represented by a collection of symbolic expressions.
Some previous works explored applications of the symbolic regression method to RL in different settings, but few considered the task of modeling environment dynamics. Our contribution consists of, first, using symbolic expressions to model the environment dynamics of the system under the influence of control inputs and, second, applying a transformer-based symbolic regression method to predict dynamics models in model-based RL setting.

\section{Related work}
In this work, we mix MBRL with symbolic regression. Therefore in this section, we discuss, first, the state of the art in MBRL, then, novel approaches in symbolic regression.

Model-based policy optimization is an approach to MBRL that optimizes the agent's policy using an environment model, also called a world model. A dynamics model is a part of an environment model that represents state transition dynamics.
Recent advances in model-based policy optimization have focused on both improving the accuracy and efficiency of dynamics models and developing new algorithms that can use these models to improve sample efficiency without loss of performance. For example, 
\cite{nagabandi2018neural} uses model predictive control in MBRL setting, while \cite{hafner2019learning} learns a world model and uses it for planning.

In \cite{janner2019trust} authors propose model-based policy optimization (MBPO) algorithm that learns an environment model and optimizes the policy in a sample-efficient manner by alternating between collecting data from the environment and optimizing the policy using the learned model. The algorithm showed state-of-the-art performance on a variety of continuous control tasks. This algorithm does not depend on the type of environment model and can be utilized for developing other policy optimization methods. Papers \parencite{hafner2019dream, hafner2020mastering} deal with the training of one-step policy with the use of a dynamics model, represented by a recurrent neural network (RNN).

In the field of symbolic regression, applications of deep learning architectures led to the development of novel methods. In \parencite{kamienny2022end} symbolic regression was modeled as a sequence-to-sequence problem and solved by using a transformer model that was trained to sequentially generate a representation of symbolic expression that would model the relationship between variables given a set of pairs of points. Previously, seq-to-seq deep learning methods had success in application to different mathematics problems, like finding indefinite integral or finding a solution to ordinary differential equation \parencite{lample2019deep}.
In \parencite{bendinelli2023controllable} the authors consider a controlled generation process that allows user-defined prior knowledge to be incorporated into the structure of the generated symbolic expression.
Symbolic expressions and methods were used in reinforcement learning in different settings. In \parencite{landajuela2021discovering} agent learns a generate a symbolic policy using an LSTM model \parencite{hochreiter1997long} and modified policy-gradient method. A similar method was used to generate symbolic expressions for solving a regression problem \parencite{petersen2019deep}.
Multiple works used genetic-programming or rule-based methods to approximate environment transition dynamics \parencite{balloch2023neuro, derner2018data}.

Our work focuses on model-based policy optimization in continuous control tasks by learning a world model using a symbolic regression method that utilizes deep learning transformer architecture.

\section{Background}
In this section, we present formal definitions and notation for MBRL and symbolic regression.
A reinforcement learning task is formulated as a Markov Decision Process (MDP). MDP is a tuple $(S, A, \tau, \rho, p_0)$, where $S$ -- set of states, $A$ -- set of actions, $\tau$ -- transition dynamics, $\rho$ -- reward function, $p_0$ -- distribution of initial states. An agent starts in an environment in some initial state $s_0 \sim p_0$.
Then it chooses action $a_t \sim \pi(s_t)$ following policy $\pi$, environment transitions to the next state $s_{t+1} \sim \tau(s_t, a_t)$, and agent observes reward $r_t = \rho(s_t, a_t, s_{t+1})$. The goal of the agent is to find optimal policy $\pi^*$ that maximizes the expected sum of rewards the agent would observe during the episode. Here a policy is a function that represents the agent's behavior. A transition is a tuple $(s_t, a_t, r_t, s_{t+1})$ that represents a unit of experience as observed by the agent. A reward function defines the task the agent will try to accomplish.

MBRL algorithms \parencite{moerland2023model} learn a model $E$ of environment dynamics $\tau$ and then use it to search for policy or value function in a sample-efficient manner, instead of trying to learn them directly from agent-environment interaction data.
In our work we use a model-based policy optimization scheme similar to \cite{janner2019trust} to train policy entirely on synthetic samples from the environment model, while concurrently optimizing both from scratch (alg. \ref{algo:mbpo}).

\begin{algorithm}
\caption{General model-based policy optimization}
\label{algo:mbpo}
\begin{algorithmic}[1]
    \algsetup{linenosize=\normalsize}
    \normalsize
    \STATE Initialize policy $\pi$, dynamics model $p_\theta$, buffer for environment samples $B_E$, buffer for model samples $B_\pi$%
    \FOR{$N$ epochs}
        \STATE sample $n$ transitions from environment under  $\pi$; add to $B_E$
        \STATE Train model $p_\theta$ on $B_E$
        \FOR{$M$ model rollouts}
            \STATE Sample $s_t$ uniformly from the last $q$ records in $B_E$%
            \STATE Perform $k$-step model rollout starting from $s_t$ using policy $\pi$; add observed transitions to $B_\pi$
        \ENDFOR
        \FOR{$G$ gradient updates}
            \STATE Update policy parameters on batch from $B_\pi$
        \ENDFOR
    \ENDFOR
\end{algorithmic}

\end{algorithm}

We also use a collection of symbolic expressions as a dynamics model. Symbolic regression is a problem of finding mathematical expression $f$ given a dataset of pairs $(x, y)$, $x \in \mathbb{R}^d$, $y \in \mathbb{R}$, such that prediction $\hat y = f(x)$ is close enough to $y$.

\section{Method}
\begin{figure}
    \centering
    \includegraphics[width=.7\textwidth]{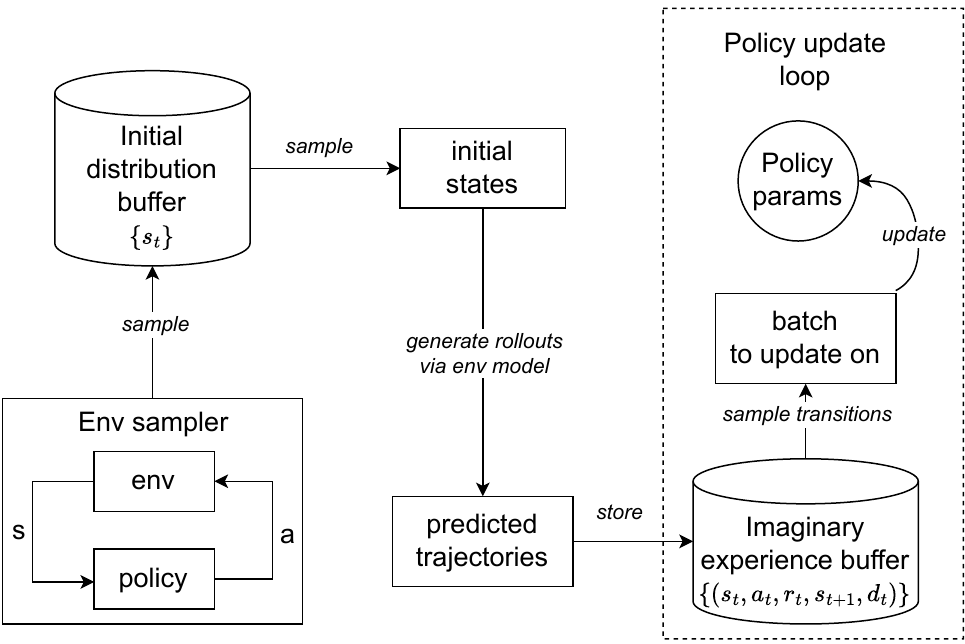}
    \caption{Sampling scheme of SAC training data. All transitions in the SAC buffer are generated using the world model. Any state that was observed during interaction with the environment is stored in the initial distribution buffer and can be sampled as the initial state for rollout generation.}
    \label{fig:sac-sampling-scheme}
\end{figure}

For policy training, we use the Soft Actor-Critic (SAC) algorithm \cite{haarnoja2018soft}, which is suitable for learning a policy over continuous action space and can utilize collected off-policy data. Policy $\pi$ is updated using data from buffer $B_\pi$, which stores transitions sampled from the environment model $E$ (fig. \ref{fig:sac-sampling-scheme}).
To generate trajectories from the environment model, we set it into the initial state and generate a short rollout of fixed length using the current policy.
Initial state distribution is a uniform distribution over the fixed-size chunk of buffer $B_E$, which contains the most recent transitions observed from the environment.
To compute reward for predicted transitions, we use true reward functions for each environment.
We use observer $g$ to map observations $o_t$ to states $s_t = g(o_t)$. Observer is implemented by a handcrafted function that maps observation from the environment to the state $s$ we model dynamics on. The observer does not use additional information and is meant to replace the encoder that would map observation to the latent variable.
We use it to simplify the task of predicting the next state by constructing exhaustively descriptive and minimal representations separately for each environment.
To successfully learn a collection of co-dependent models, it's important to balance their sampling and update rates. We fix these as hyperparameters.

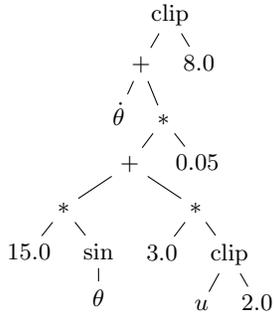
\begin{figure}
\centering
\footnotesize
\begin{forest}
for tree={
    l=0pt,
    l sep=5pt
}
[$\operatorname{clip}$ [$+$ [$\dot{\theta}$ ][$*$ [$+$ [$*$ [$15.0$ ][$\sin$ [$\theta$ ]]][$*$ [$3.0$ ][$\operatorname{clip}$ [$u$ ][$2.0$ ]]]][$0.05$ ]]][$8.0$ ]]
\end{forest}
\caption{Tree representation of the expression $\operatorname{clip}((\dot{\theta} + (((15.0 * \sin(\theta)) + (3.0 * \operatorname{clip}(u, 2.0))) * 0.05)), 8.0)$ for the angular velocity of the pendulum, generated by transformer model that was trained on pendulum dynamics.}
\label{fig:expression-tree-pendulum-angvel}
\end{figure}

We build a model of environment dynamics using an ordered set of symbolic expressions that collectively map the current state and an action to the next state by $\tau\colon (s, a) \mapsto s'$, where each expression determines one coordinate of $s'$. These expressions are generated by the transformer model \parencite{kamienny2022end} that was trained to generate expression $f$ given a set of pairs of points $\{(x, y)\}$ such that prediction $\hat{y} = f(x)$ is close to the target $y$. Each expression is generated as a sequence of tokens that represent elementary functions, predefined constants, or real numbers. This sequence is arranged into a tree that corresponds to a mathematical expression. For example, the generated expression for angular velocity in the Pendulum environment is shown in fig.~\ref{fig:expression-tree-pendulum-angvel}.

We train the transformer model in a supervised regime beforehand and do not update its parameters during policy learning. In our experiments, we use two kinds of model parameters. Parameters of the first kind were provided by the authors of \parencite{kamienny2022end}, who trained the model on a synthetic dataset of randomly generated elementary functions. Parameters of the second kind we obtain by training the transformer model on environment-specific datasets, collected by different exploration policies, like uniform policies with fixed action hold periods. For each environment, we train a separate transformer model. During the policy learning, the transformer model is periodically inferred, mapping observed transitions to a set of candidate symbolic expressions that describe the transitions. Next, the parameters of each candidate are optimized using BFGS method \parencite{fletcher2000practical} on the same dataset, candidates are ranked by prediction error, and the best candidate is chosen. This symbolic expression is added to the collection that is used as an environment model to generate transitions for policy learning.
If the symbolic expression cannot be evaluated at some points, we regenerate it and do not mask any tokens, like \texttt{sqrt}.
All trajectories observed from an environment are stored in buffer $B_{EM}$. We uniformly sample batches of transitions from this buffer to infer the transformer model at them.

Our method differs from previous works in the choice of dynamics model and training schedule.
We use a collection of symbolic expressions to model environment dynamics and evaluate our method only in environments with vector observations without the need to solve the problem of encoding input image observations. For some environments, we use a handcrafted function to map observations to other quantities that are more favorable to model dynamics on. These functions have no trainable parameters and do not affect the policy input.

\section{Interpretability}

The field of interpretable RL studies different aspects of RL, such as interpretable inputs, interpretable policies \parencite{kenny2022towards, hein2020interpretable, verma2018programmatically, mott2019towards}, and interpretable world models \parencite{glanois2021survey}. The symbolic form of expressions we use is more suitable for analysis than neural networks. For policies, such analysis may be conducted to make sure safety constraints are satisfied. For transition models, interpretable representation allows one to apply other control methods that may require it or to draw additional insights about the environment that may help in control system design. Since an interpretable symbolic representation usually corresponds to a much simpler function than a trained neural network does, a symbolic transition model may exactly capture the dynamics of a physical environment, leading to better extrapolation quality.

We observe a structural similarity between generated expressions and true dynamics only in cases where the transformer model was trained on the dynamics of the same environment it generates expressions for. For example, with a transformer model trained on a dataset of random functions, the symbolic regression method we use generates the following expression for the angular velocity in the Pendulum environment: $(((7.99 * \dot{\theta}) + (0.0535 + (0.297 * u))) + ((1.82 + (-0.0089 * u)) * (-0.0297 + (-0.411 * \cos((((-0.0744 + (3.153 * \theta)) + (-0.0749 + (-0.331 * \sqrt{((-3*10^{-5} + (2.58 * \sin((1.22 + (0.0273 * \theta))))))}))) - (-8.52 + (-0.0233 * \dot{\theta}))))))))$, which accounts for data standardization to $[-1, 1]$. Compared to the true dynamics in fig.~\ref{fig:expression-tree-pendulum-angvel}, this expression does not match it, since the model did not have a token for the $\operatorname{clip}$ function in its dictionary. Though this expression may not provide any fruitful insights for other purposes, its prediction error is low enough for successful learning, since the policy optimization method we use generates only short rollouts.

\section{Experiments}

\begin{figure}[htb]%
\begin{subfigure}{0.22\textwidth}
    \centering
    \includegraphics[width=1\columnwidth]{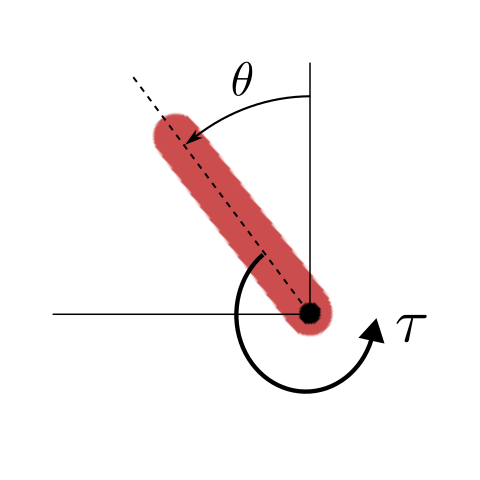}
    \caption{Pendulum}
    \label{fig:frame-pendulum}
\end{subfigure}%
\hfill
\begin{subfigure}{0.22\textwidth}
    \centering
    \includegraphics[width=1\columnwidth]{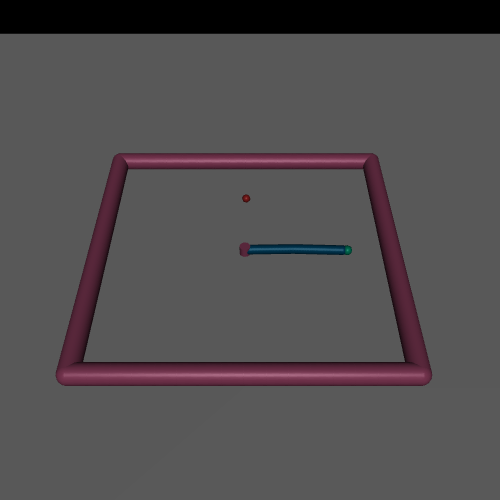}
    \caption{Reacher}
    \label{fig:frame-reacher}
\end{subfigure}%
\hfill
\begin{subfigure}{0.22\textwidth}
    \centering
    \raisebox{12pt}{
    \includegraphics[width=1\columnwidth]{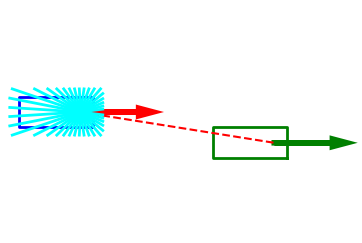}
    }
    \caption{Car2d}
    \label{fig:frame-car2d}
\end{subfigure}%
\caption{Illustrations of environments. The Pendulum environment (\subref{fig:frame-pendulum}) represents the task of balancing a swinging pendulum in an upright position, the agent observes angle $\theta$ and angular velocity and controls the torque applied at the hinge. Reacher (\subref{fig:frame-reacher}) represents the task of reaching a target point by the tip of the two-link planar manipulator, agent observes joint positions and velocities, target point location, vector between the target and the manipulator's tip, and controls the torques applied at the joints. Car2d (\subref{fig:frame-car2d}) represents the task of parking at the target location a car that can move forward and steer in a plane, the agent observes position, orientation, velocity, steer, and target location, and controls acceleration and angular velocity of steering.}
\end{figure}

We evaluate our method, SAC \parencite{haarnoja2018soft}, MBPO \parencite{janner2019trust}, and Dreamer-v3 ( ``medium'' configuration of hyperparameters) \parencite{hafner2023mastering} in several simulation environments and compare their performance. Environments we use are based on Pendulum (fig.~\ref{fig:frame-pendulum}) and Reacher-v2 (fig.~\ref{fig:frame-reacher}) from Gym \parencite{brockman2016openai}, as well as a custom environment for controlling a car on a plane (fig.~\ref{fig:frame-car2d}). They are built using the MuJoCo simulator. Our version of Reacher allows to control joint positions through hand-designed PD regulators to facilitate learning of an environment model and has adjusted a reward function that does not include control cost. We limit an episode length to $200$ steps for the Pendulum, $50$ steps for Reacher, and $250$ steps for the Car2d environment. The experimental results are averaged over 3 runs, the shaded area shows the standard deviation.

The transformer model operated on a set of tokens that were able to represent the following elementary functions and constants: addition, subtraction, multiplication, division, absolute value, inverse, square root, logarithm, exponentiation, sine, cosine, arcsine, arccosine, tangent, arctangent, clipping, rounding towards zero, $e$, and $\pi$.

\begin{figure}[htb]
    \centering
    \begin{subfigure}[b]{0.5\textwidth}
        \centering
        \includegraphics[width=1.\textwidth]{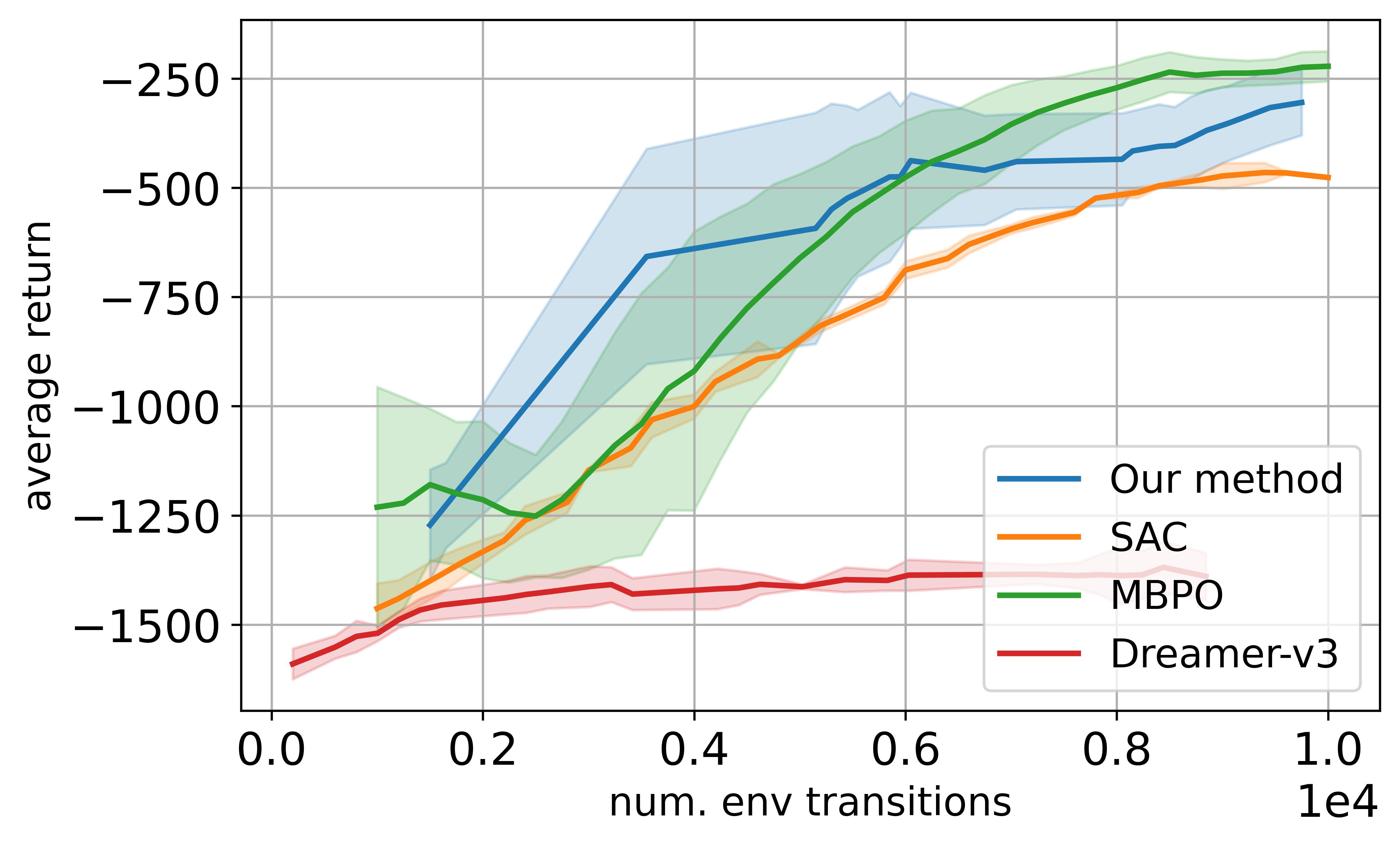}
        \caption{Average return}
        \label{fig:mbpo-pendulum-learning-curves}
    \end{subfigure}%
    \begin{subfigure}[b]{0.5\textwidth}
        \centering
        \raisebox{0pt}{
        \includegraphics[width=.85\linewidth]{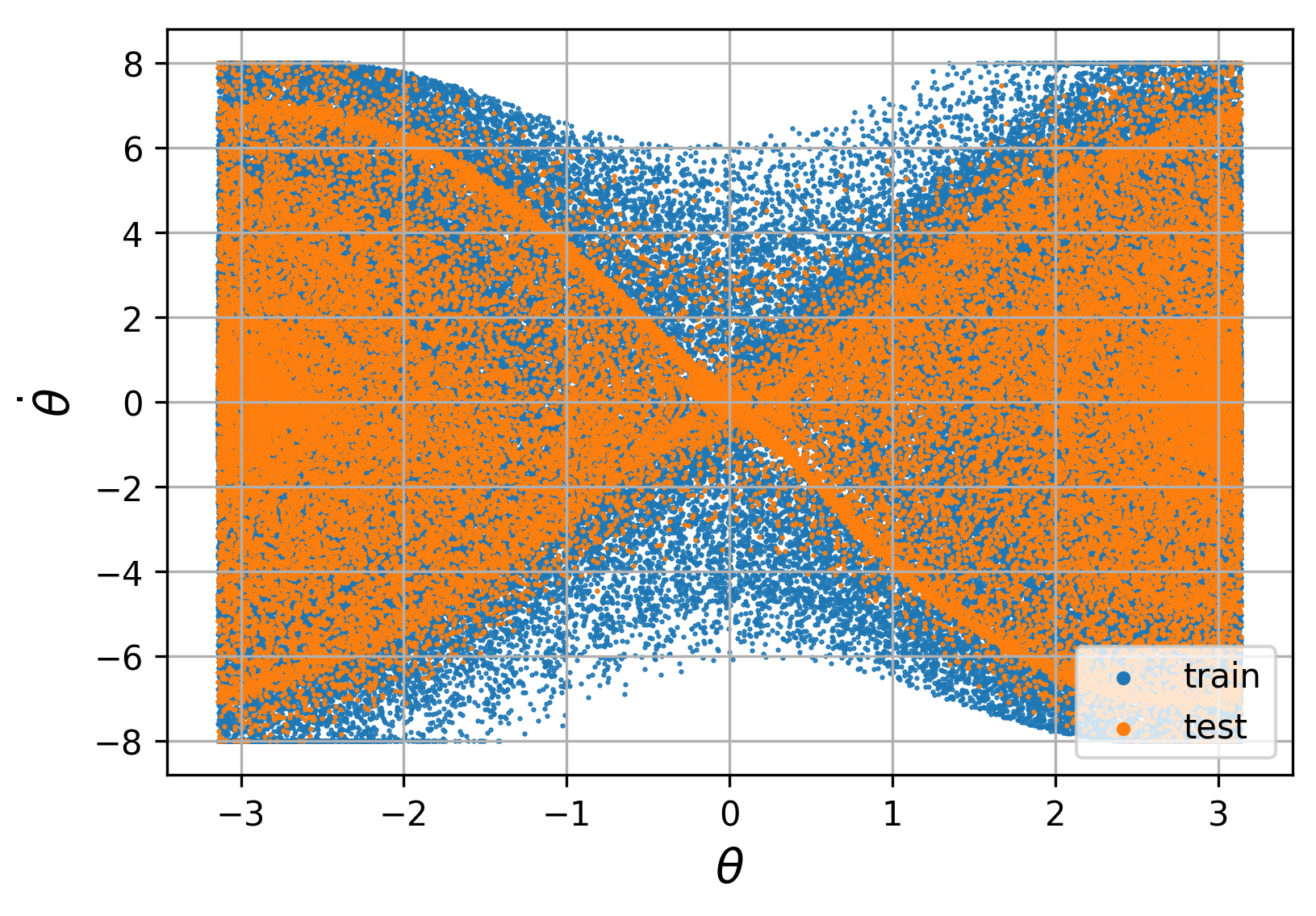}}
        \caption{State space coverage}
        \label{fig:dm-input-space-coverage}
    \end{subfigure}%
    \caption{Method evaluation in the Pendulum environment. State space coverage shown for the collected dataset (train) and samples from joint policy-environment distributions (test) during agent training.}
\end{figure}

We evaluate our method in the Pendulum environment in multiple settings. In all of them, we map observation from $(\cos(\theta), \sin(\theta), \dot{\theta})$ to $(\theta, \dot{\theta})$ so that transition model did not have to learn to invert trigonometric functions as output depends directly on $\theta$. First, we run our method using general-purpose weights for the transformer model that were not trained on pendulum dynamics. Our method outperforms SAC and Dreamer-v3 in terms of sample efficiency (fig.~\ref{fig:mbpo-pendulum-learning-curves}). Dreamer-v3 exceeds the average episode return value of $-250$ after approximately $4\mathrm{e}4$ observed environment transitions.
Second, we train a transformer model to predict the dynamics of the Pendulum environment and then use it in agent training runs.
We collect $5000$ trajectories of $50$ transitions each using a set of $10$ policies that sample an action from uniform distribution and hold it for $k \in [1..10]$ steps, where $k$ is fixed for each policy.
Since transformer parameters are not updated during the MBRL run, we ensure that during supervised training of the transformer, the dataset is representative of all distributions induced by MDP and a policy as it is updated during the run (fig.~\ref{fig:dm-input-space-coverage}).

\begin{figure}[htb]
    \centering
    \includegraphics[width=.65\linewidth]{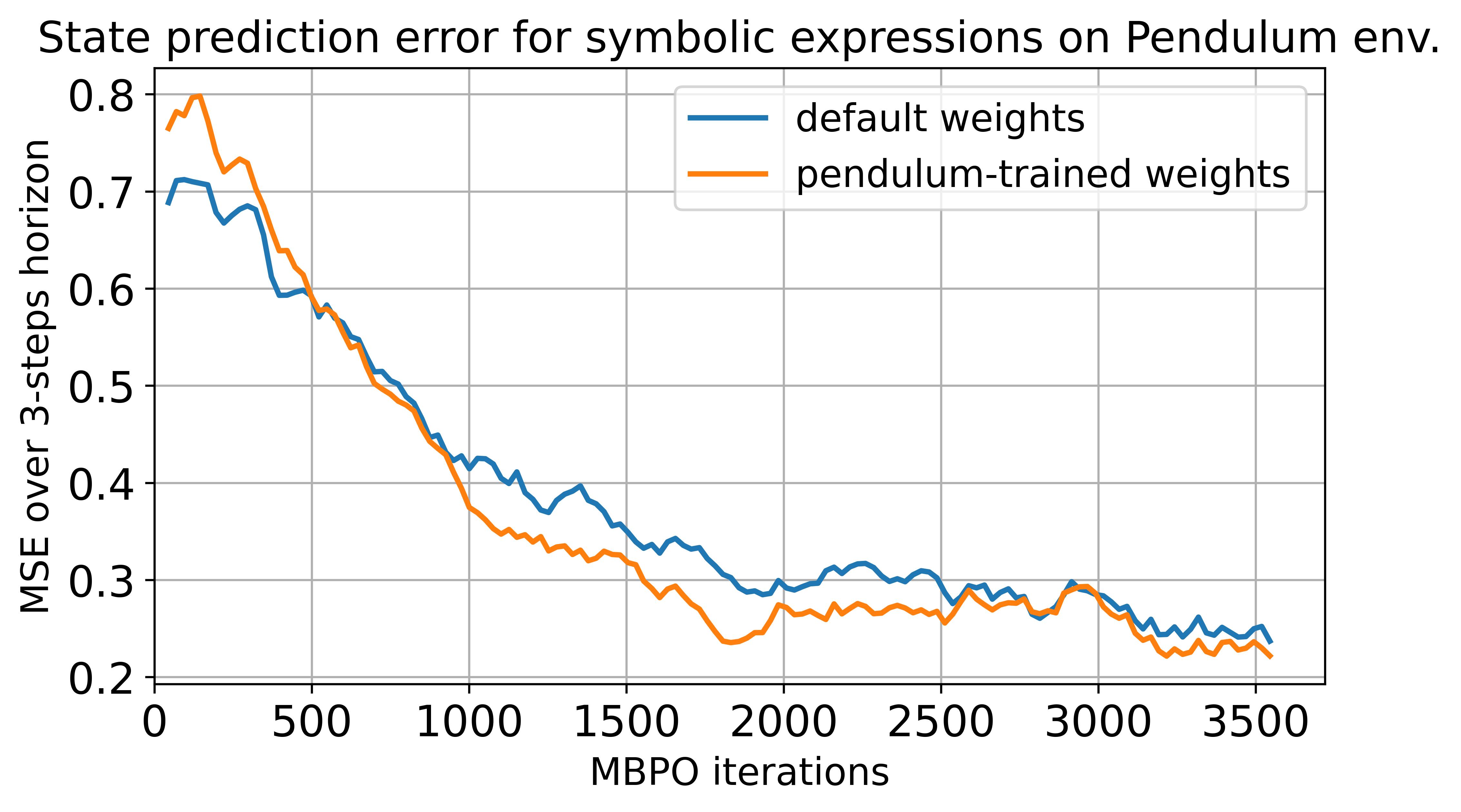}
    \caption{3-step MSE of predicted trajectories in the Pendulum environment.}
    \label{fig:xfmr-pretraining-pendulum}
\end{figure}

The trained transformer model can predict the target expression exactly. For example, for the angle of the pendulum $\theta$ the output is $\theta + (\operatorname{clip}((\dot{\theta} + (((15.0 * \sin(\theta)) + (3.0 * \operatorname{clip}(u, 2.0))) * 0.05)), 8.0) * 0.05)$, where $\dot{\theta}$ is angular velocity and $u$ is applied torque. This pre-trained model did not improve the agent's learning curves, but greatly increased prediction quality for the longer horizons (fig.~\ref{fig:xfmr-pretraining-pendulum}).
The absence of improvement in RL runs is probably because our method samples only short trajectories of 1 or 2 transitions from the ground truth states and does not benefit from increased prediction quality.
\begin{figure}[htb]
    \centering
    \begin{subfigure}[c]{0.5\textwidth}
        \includegraphics[width=1.\linewidth]{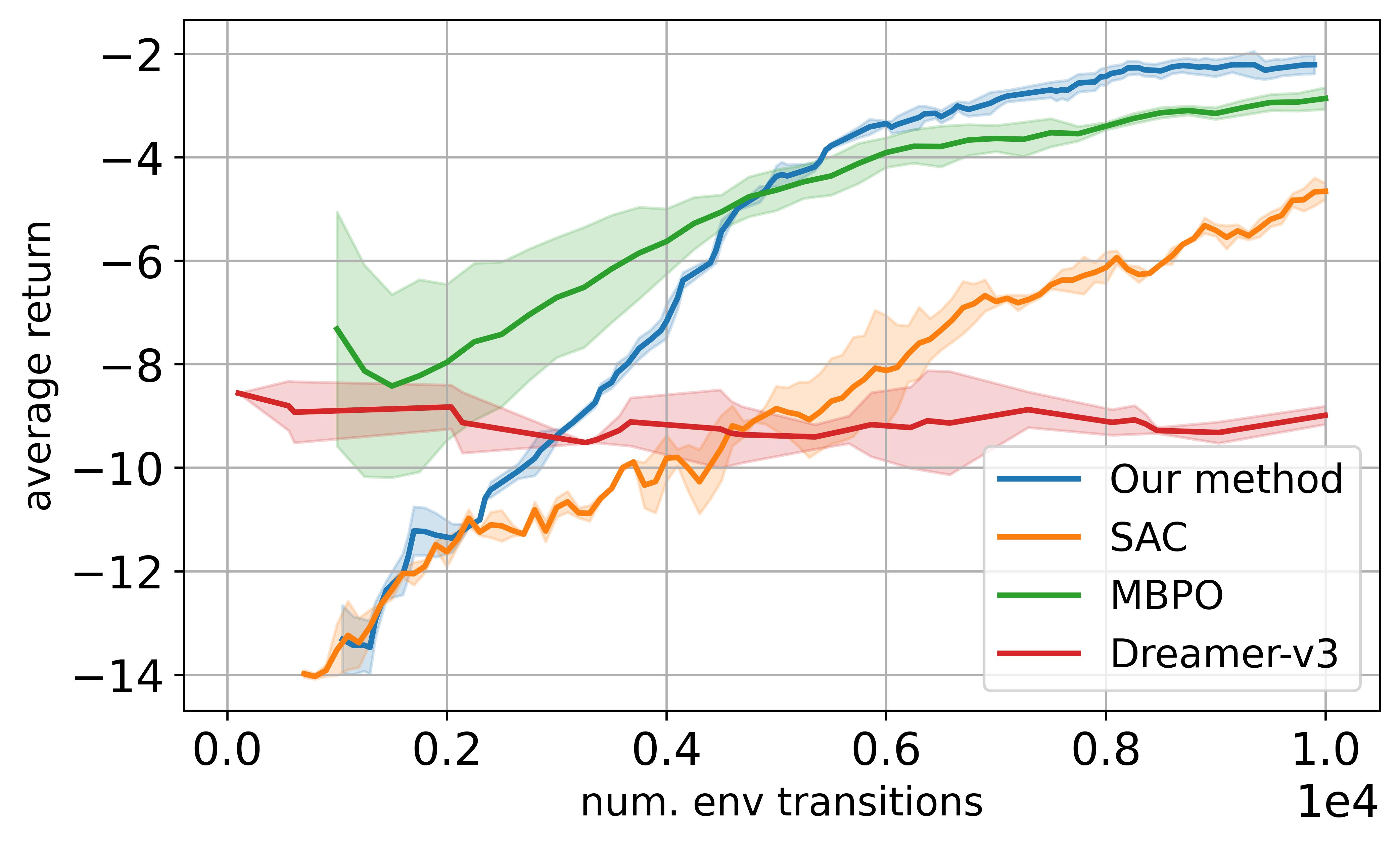}
        \caption{Average return in the Reacher environment}
        \label{fig:mbpo-reacher-learning-curves}
    \end{subfigure}%
    \begin{subfigure}[c]{0.5\textwidth}
        \centering
        \includegraphics[width=1.\linewidth]{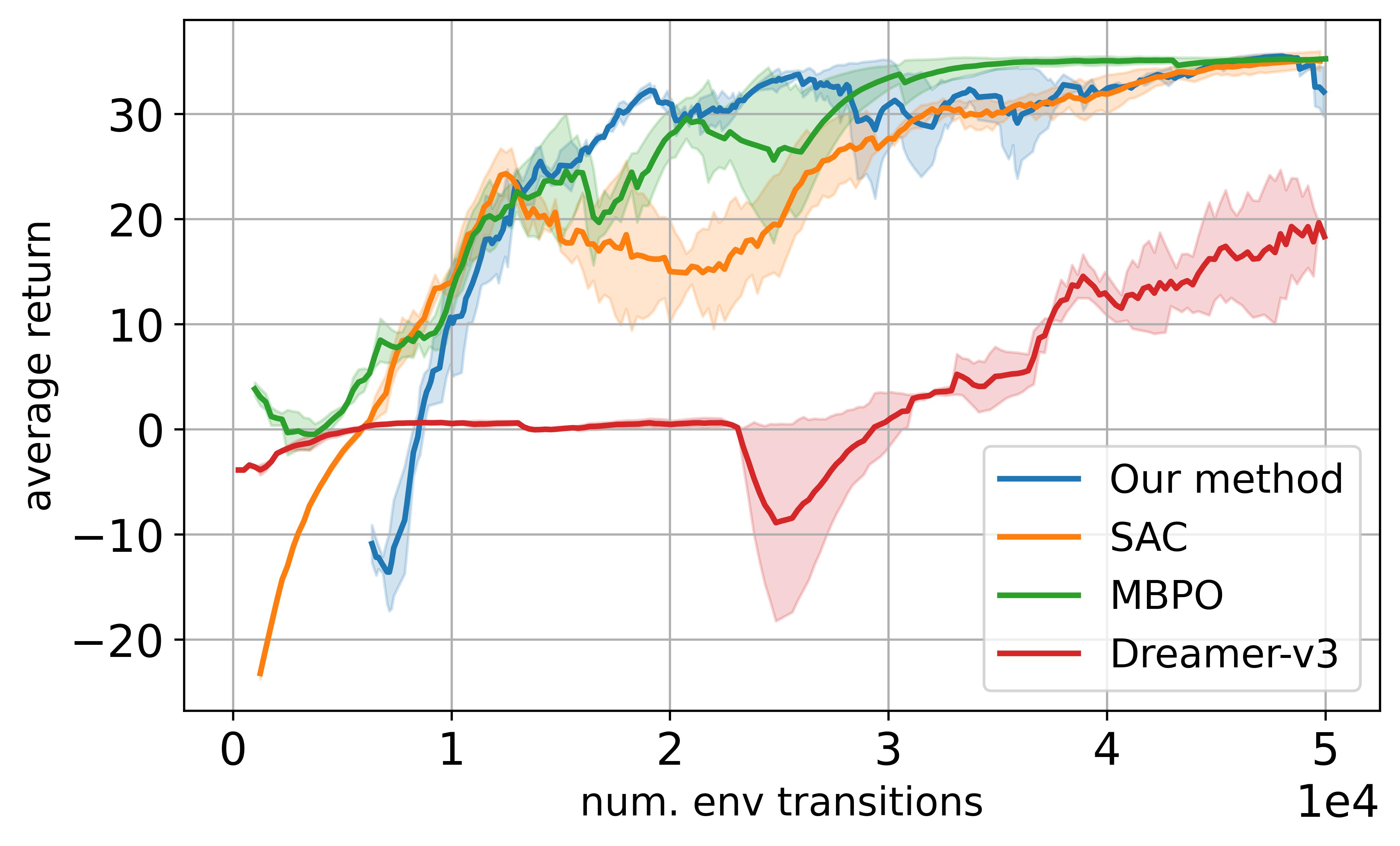}
        \caption{Average return in the Car2d environment}
        \label{fig:exp-card2}
    \end{subfigure}%
    \caption{Method evaluation in the Reacher and Car2d environments. In Car2d environment model-free SAC reproducibly experiences performance drop after it learned to move the car to the target location with an overshoot. Since the car cannot drive backward, it has to drive another lap.}
\end{figure}

We evaluate our MBRL method in Reacher using the same dynamics model configuration (fig. \ref{fig:mbpo-reacher-learning-curves}).
Results show that our method outperforms SAC and Dreamer-v3 in terms of sample efficiency, as expected. Dreamer-v3 reaches the average episode return value of $-3.7$ after approximately $1\mathrm{e}5$ observed environment transitions.

We evaluate our method in the Car2d environment, where the task is to navigate a car to the target location. Our method shows asymptotic performance and sample efficiency similar to SAC and MBPO and learns faster than Dreamer-v3 (fig.~\ref{fig:exp-card2}). Dreamer-v3 exceeds the average episode return value of $30$ after approximately $1.5\mathrm{e}5$ observed environment transitions.

\begin{figure}[htb]
    \centering
    \includegraphics[width=0.55\linewidth]{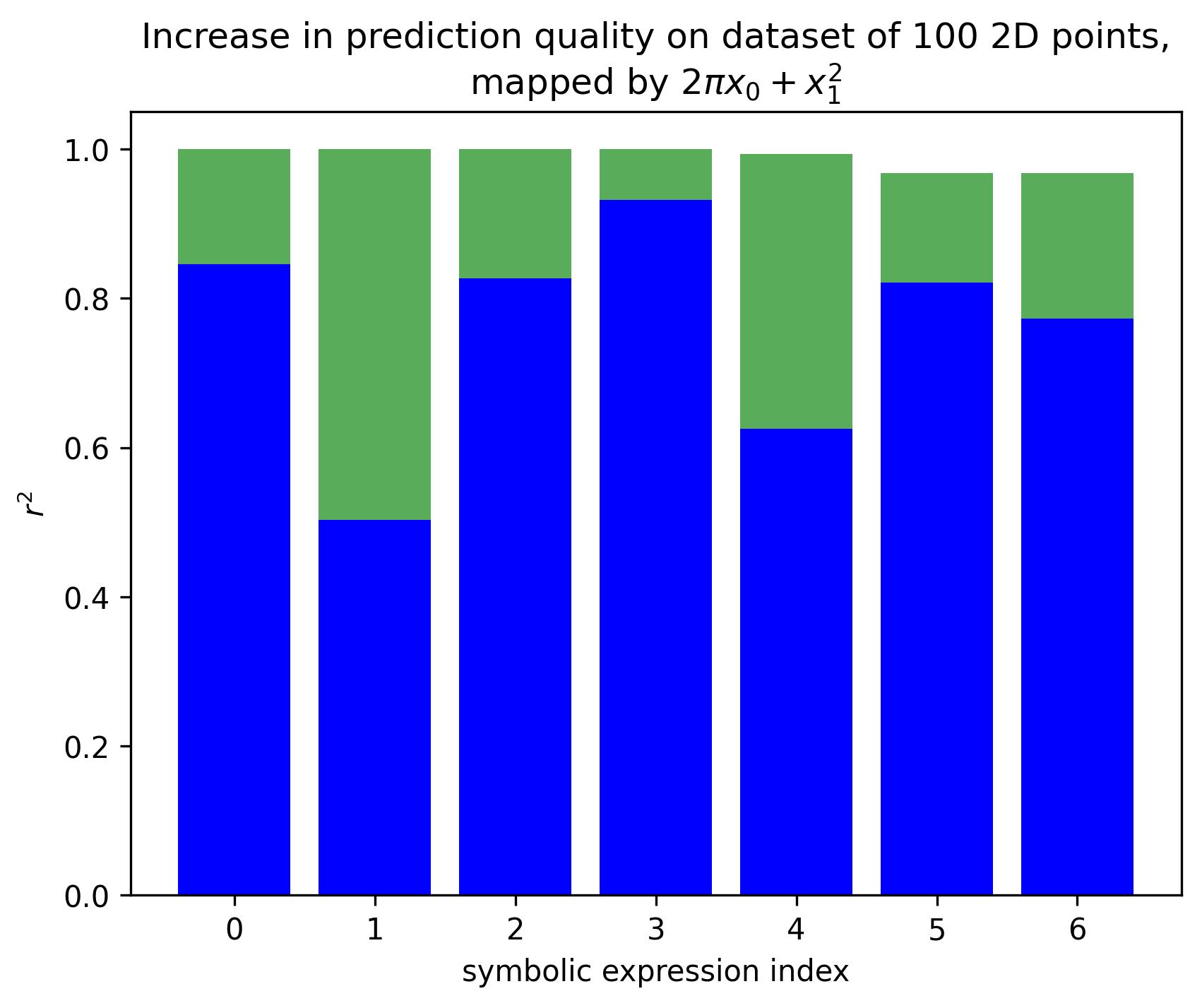}
    \caption{Increase in $r^2$ of prediction quality of $7$ candidate symbolic expressions (shown on the $x$-axis by their index), generated by transformer model, after optimization by BFGS (shown as green).}
    \label{fig:bfgs-impact}
\end{figure}

We evaluate the impact of optimizing the parameters of a symbolic expression by BFGS algorithm after it was generated by a transformer model. On a simple dataset that consists of $100$ points $x \in \mathbb{R}^2$ and their images under $x \mapsto 2 \pi x_0 + x_1^2$, we observe a significant increase in prediction quality, measured by the coefficient of determination $r^2$, around $0.2$ (fig.~\ref{fig:bfgs-impact}).

\section{Conclusion}

This paper presents a method for model-based policy optimization that uses a symbolic regression method based on transformer architecture to infer a set of symbolic expressions representing an environment's dynamics model. We evaluate the proposed method on different robotic control tasks and compare it to other model-based algorithms and a model-free SAC algorithm.

The proposed method has several limitations. First, it cannot handle high-dimensional problems that cannot be decomposed into simpler problems, like control of a 7-DoF arm or humanoid robot. Second, the transformer model can account only for a limited number of observed points due to inference complexity. Third, the generated world model cannot be incrementally updated and does not depend on the previous inference result, which can lead to instant degradation in prediction quality. Fourth, the transformer model is inferred separately for each output coordinate, while their expressions may be mostly similar.

Future development of this approach may include conditioning the inference process on the previous result, reusing and sharing parts of similar expressions between all coordinates during inference, factorizing an environment's dynamics into parts suitable for the SR method, and handling the rest of the dynamics using other methods.

\printbibliography

\end{document}